\documentclass[letterpaper, 10 pt, conference]{ieeeconf}  

\IEEEoverridecommandlockouts                              

\usepackage{cite}
\usepackage{graphicx}
\usepackage{amsmath,amsfonts,amssymb}

\title{\LARGE \bf
Egocentric Visual Self-Modeling for Autonomous Robot Dynamics Prediction and Adaptation
}

\author{Yuhang Hu* \quad Boyuan Chen \quad Hod Lipson\\
Columbia University, Duke University
\thanks{This work was supported in part by the US National Science Foundation (NSF) AI Institute for Dynamical Systems (DynamicsAI.org), grant 2112085. *yuhang.hu@columbia.edu.}%
}

\begin{document}

\maketitle
\thispagestyle{empty}
\pagestyle{empty}

\begin{abstract}
The ability of robots to model their own dynamics is key to autonomous planning and learning, as well as for autonomous damage detection and recovery. Traditionally, dynamic models are pre-programmed or learned from external observations. Here, we demonstrate for the first time how a task-agnostic dynamic self-model can be learned using only a single first-person-view camera in a self-supervised manner, without any prior knowledge of robot morphology, kinematics, or task. Through experiments on a 12-DoF robot, we demonstrate the capabilities of the model in basic locomotion tasks using visual input. Notably, the robot can autonomously detect anomalies, such as damaged components, and adapt its behavior, showcasing resilience in dynamic environments. Furthermore, the model's generalizability was validated across robots with different configurations, emphasizing its potential as a universal tool for diverse robotic systems. The egocentric visual self-model proposed in our work paves the way for more autonomous, adaptable, and resilient robotic systems.
\end{abstract}

\section{Introduction}
\label{sec:intro}
Human vision plays a pivotal role in facilitating a wide range of agile behaviors. The visual feedback it provides is instrumental in tasks ranging from basic locomotion to complex motor skills\cite{pearson1993common,alexander2003principles,marigold2008visual}. In contrast, individuals with visual impairments often face challenges in performing these behaviors due to the absence of this feedback. This reliance on vision is not unique to humans. Many sighted animals display more sophisticated and agile locomotion patterns than their counterparts with limited visual capabilities \cite{gibson1958visually, marigold2007gaze}. This observation emphasizes the importance of vision in locomotion and suggests that robots, especially legged robots, could benefit significantly from visual feedback.

Traditionally, legged robot locomotion has relied on proprioceptive sensors, such as IMUs and joint encoders, with vision being used primarily for high-level planning tasks. This is due to the relatively low frequency of most visual perception systems (less than 30 Hz) compared to the high control frequencies required for stable locomotion (often exceeding 200 Hz)\cite{yu2021visual, yang2021learning,loquercio2023learning}. While some works have explored the use of multimodal sensory input, combining vision and proprioception to train locomotion policies using reinforcement learning\cite{fu2022coupling,yang2021learning}, the potential of purely visual observation for legged robot locomotion remains unexplored. Therefore, investigating this potential could provide a foundation for combining vision into the locomotion controller, leading to more adaptable and resilient legged robots.

Self-modeling is a form of model-based control that decouples (deconvolves) a model of the robot (the ``self-model'') from a model of its environment and task\cite{nguyen2011model,bongard2006resilient}. The key idea is that while the task and environment of a robot may change frequently, the robot itself is relatively constant across different tasks and environments. Therefore, separating out the self-model from the model of everything else and reusing the self-model for new tasks, can simplify adaptation in situations that involve lifelong learning under variable tasks. 

For example, if a tennis player must learn how to play badminton, they may need to learn new game rules, but will not need to relearn how to run across a court or grasp a racquet, because running and grasping are common actions performed by the same body. In this way, self-modeling facilitates transfer learning \cite{yosinski2014transferable}. 

Importantly, the value of a self-model as compared to model-free reinforcement learning increases with the complexity of the robot \cite{kwiatkowski2022deep,hu2023scalable} and the complexity of the task. Therefore, finding efficient ways to create accurate self-models is extremely valuable and will offer a notable practical advantage as robot complexity continues to increase.

In this paper, we explore the potential of leveraging purely visual data into robotic systems to achieve locomotion and damage resilience capabilities. Our approach goes beyond merely using vision as a pose estimator. we aim to enable robots to model their own dynamics, kinematics, and morphology using visual data instead of using explicit equations, CAD models, and information from IMUs and external optical trackers which requires extensive prior knowledge or infrastructure that may not exist when robots must learn to adapt to new conditions in-situ.

\begin{figure*}[t]
    \centering
    \includegraphics[width=0.8\textwidth]{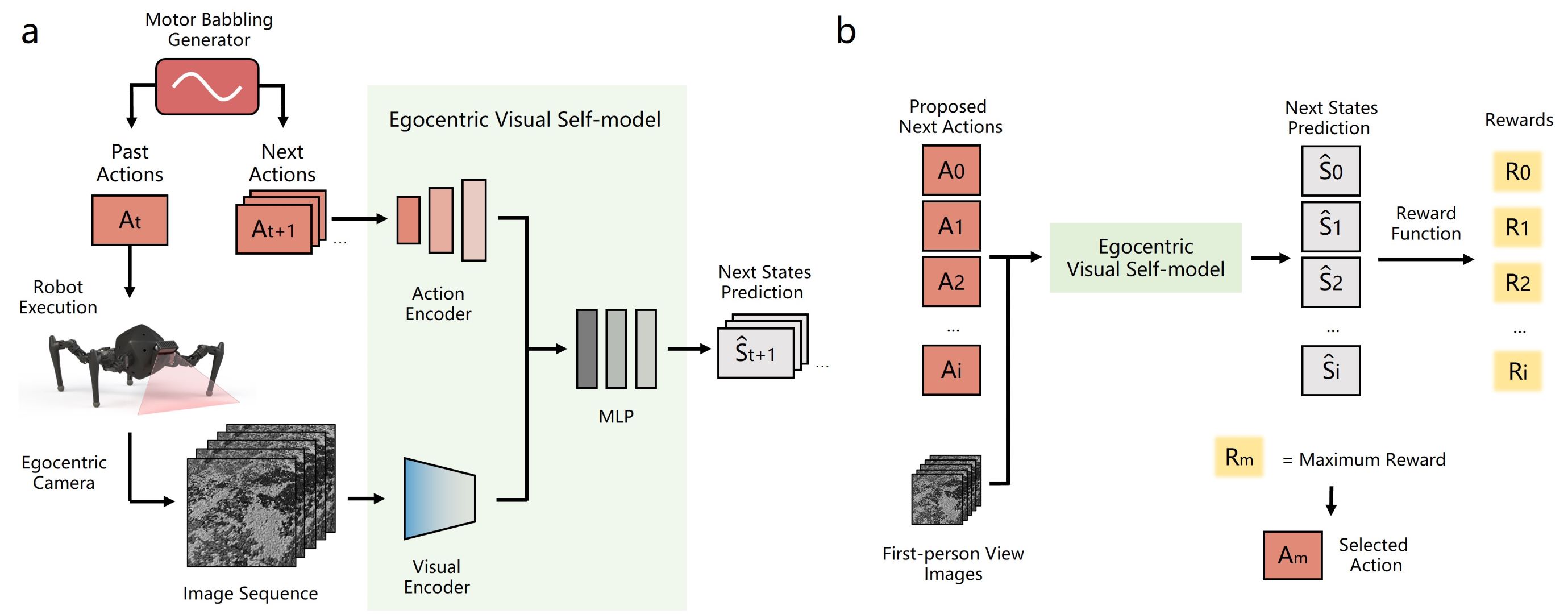}
    \caption{\textbf{Egocentric Visual Self-Model. a) Training pipeline.} Motor babbling generates random actions $A_{t}$ executed by the robot. The onboard camera captures sequential images processed through a visual encoder. Proposed future actions $A_{t+1}$ are encoded and concatenated with visual features to predict the next robot state $\hat{S}_{t+1}$. \textbf{b) Deployment.} For real-time control, multiple proposed actions $A{0...i}$ are input to the trained self-model along with current visual data. Based on these predictions, rewards $R_{0...i}$ are computed for each predicted state$S_{0...i}$. The robot then selects the action $A_{m}$associated with the highest reward $R_{m}$ for the next move.}
    \vspace{-10pt}
    \label{fig:overview}
\end{figure*}

Our work presents how a robot can learn to predict its future state, conditioned on motor commands, using a sequence of frames from a single RGB camera (Fig. \ref{fig:overview}). Our proposed egocentric visual self-model can be used in several ways. For example, a planning algorithm can use the model to test the consequence of various candidate motor commands and use these predictions to choose the action most suited for a particular task - before executing any of these tasks in reality. Alternatively, a damage detector can use this model to compare predicted motion to actual motion (from visual optometry). If predicted motion deviates from actual motion, something must have gone wrong. These applications and other examples will be demonstrated in this paper. We believe that this approach can be extended, allowing robots to exhibit more agile and adaptive behaviors. 

Although it is no longer a challenge for a legged robot to walk on flat ground in 2024, our work is to provide evidence of the potential of vision in walking tasks to catalyze further development of more dynamic and adaptable robotic systems integrating visual perception with robotic gait control. Our key contributions can be summarized as follows:

1. This work demonstrates a legged robot successfully performing locomotion tasks using purely egocentric visual observations without relying on proprioceptive sensors. Our visual egocentric self-model network effectively fuses sequences of high-dimensional visual observations and actions to predict future robot states

2. A domain randomization and data augmentation pipeline for robot locomotion that enables effective zero-shot sim-to-real transfer.

3. Validation of the generalizability of our approach across multiple robots with varying morphologies and complexities, highlighting its potential as a universal tool for diverse robotic systems.

4. A demonstration of the ability to recover a damaged robot's locomotion capabilities by re-training the self-model using purely visual perception data collected in the real world, showcasing the adaptability and resilience of our approach in the face of unexpected challenges.

\section{Related works}
\label{sec:related_works}

\textbf{Robot Locomotion.} Historically, most legged robots have been designed with a "blind" approach, relying primarily on pre-programmed kinematics or dynamics with low dimensional sensor feedback for locomotion tasks\cite{carpentier2021recent, torres2022legged,maroger2020walking, peng2020learning, gangapurwala2020guided}. However, with the advent of advanced computer vision and machine learning techniques, there has been a paradigm shift in the design and control of legged robots. Recent works have started to integrate vision into legged robot systems, albeit often as a separate module. For instance, the Cheetah 3 robot developed at MIT uses vision to navigate complex terrains and obstacles, but the visual system operates largely independently from the robot's control algorithms \cite{bledt2018cheetah}. Similarly, the ANYmal robot, designed by Robotic Systems Lab, employs visual perception for tasks like mapping and localization, but the semantic information extracted from visual data is primarily used for navigation rather than influencing the robot's intrinsic dynamics \cite{hutter2016anymal}. Spot by Boston Dynamics is another example where vision aids in navigation and object manipulation, but the robot's dynamics are primarily determined by pre-programmed algorithms and sensors \cite{bouman2020autonomous}. This design philosophy, using vision as a separate module, contrasts starkly with the natural world, where sighted beings, from humans to animals, utilize their visual perception to navigate complex terrains, avoid obstacles, and perform intricate tasks with remarkable agility. Our proprioceptive senses are, by themselves, insufficient for controlling our locomotion ability\cite{loomis1993nonvisual,kallie2007variability, gibson2014ecological}. Most of us rely on visual feedback to provide signals for sustained locomotion. Vision not only helps us navigate and avoid obstacles but also helps us keep balance and fine-tune our own kinematic self-awareness\cite{warren2001optic,paulus1984visual}. For example, if we see patterns on the ground moving backward, our brain can implicitly deduce that we are moving forward. In addition, our brain can learn to predict what will be the effect of certain muscle actions -- in other words, our brain uses visual feedback to create and maintain our predictive dynamic self-model\cite{wolpert2001motor,shadmehr2008computational}. 

\textbf{Self-modeling.} The use of self-models for robot control is not new. Almost all robots today use some form of model predictive control\cite{katz2019mini,lenz2015deepmpc,salzmann2023real}. These forward models are either analytical or data-driven. Analytical self-models are programmed using equations, or using simulated physics and CAD geometries \cite{todorov2012mujoco,coumans20162019,1389727}. Data-driven models are learned using data captured from inertial measurement units (IMUs) and/or external tracking devices such as cameras and optical trackers \cite{siekmann2021blind,rudin2022learning}. 

Data-driven self-modeling acquires and maintains a computational representation of the robot morphology, kinematics and dynamics through a machine-learning process that is typically bootstrapped during an initial period of safe ``motor-babbling'', and potentially continued over the lifetime of the machine. Past work involved learning self-models using data from internal inertial sensors or external cameras \cite{chen2021full, kwiatkowski2019task,hu2023teaching}. Each of these self-modeling approaches however entailed its own limitation: Analytical models account only for limited dynamics. Data driven models that rely on proprioceptive sensors (e.g. IMUs) drift with time. Self-models that use external observations (e.g. external cameras) are not portable. Our goal here is to create self-models without proprioceptive or external sensors, and without relying on provided equations and existing geometric models. 



\section{Methodology}

The robot observation space consists solely of an egocentric video feed (Fig. \ref{fig:overview}). By combining this visual data with action commands, the Egocentric Visual Self-Model can predict the robot's future state. We assume that the visual data contains rich information about the ground, such as surface and texture, which provides motion cues when viewed as a sequence of frames. Moreover, visual modeling approaches benefit from recent advancements in computer graphics, where modern simulators offer realistic image renderings that enable effective pre-training in simulation and sim-to-real transfer \cite{staranowicz2011survey,andrychowicz2020learning, akkaya2019solving}. The self-model learns the computational dynamics of the robot body in a self-supervised manner. During modeling, the self-model takes a sequence of egocentric images and motor commands to predict the robot's future states (Fig. \ref{fig:overview}a). This process allows the model to implicitly learn properties such as mass distribution, geometry, friction, actuation, and sensing delays. Once acquired, the self-model can be used by any Model Predictive Control algorithm in real-time to satisfy various objectives (Fig. \ref{fig:overview}b).

\subsection{Data Representation}

At each timestep $t_n$, the commands $A_t$ execute over a fixed period, during which the camera captures five consecutive images $I_{t,0}-I_{t,4}$. Instead of directly using the camera readings, we crop the input RGB images from $320 \times 240 \times 3$ to $240 \times 240 \times 3$ and resize them to $128 \times 128 \times 3$. Furthermore, we only use the gray-scale image rather than the raw RGB input to reduce the dimension of the visual observation by a large margin. Such a process is critical for real-time control of the physical robot. This is because all program steps are serial. By having a smaller network to process the reduced visual inputs, we can save the cost of the decision-making process during inference time, thus avoiding losing too much information about the robot states while the robot moves constantly. 

In addition to the egocentric visual observations, the robot also takes its next-step action as inputs. Since we use position control for the robot, the action commands are encoded as a vector with 12 numbers representing the angular displacement for each motor. Our egocentric visual self-model aims to predict the robot states from its visual observation and next-step action. However, since we do not rely on any GPS-like systems to localize the robot, we do not have the pose of the robot in the global world coordinate. We therefore define the robot state as the change of the robot position and orientation in the robot coordinate frame known as  $\Delta x,\Delta y, \Delta z, \Delta roll, \Delta pitch, \Delta yaw$.

\subsection{Data Collection and Augmentation}

We collect data in the Pybullet robotics physics simulation \cite{coumans20162019}. Initially, we used a random policy that uniformly samples actions in a normalized action space [-1, 1] and recorded images and robot state. Our gait generator provides a safe and effective exploration range to generate diverse gaits and prevent meaningless or hazardous actions. It is crucial when dealing with real-world robotic applications where safety and efficiency are essential. However, gait generators alone do not provide high-quality locomotion policies, as shown in all our baseline comparisons where the gait generators are applied uniformly in the experiments section. The details of the gait generator are included in the supplementary materials.

To bridge the gap between simulation and the real world, we randomize three aspects of the simulated environment: observation, action, and friction. We also apply data augmentation during training, such as randomly cropping, rotating, and translating the rug texture images, and adding brightness and Gaussian noise to the observation images. These techniques force the model to learn the relationships among the five frames rather than the patterns within individual images. Additionally, we add noise to the robot joint positions and torques and randomize the friction coefficient during data collection to simulate real-world conditions.

\subsection{Architecture and usage of the model}

The architecture of the egocentric visual-self model is implemented as a hybrid of residual networks and recurrent neural networks \cite{he2016deep}. The model takes two pieces of information: a sequence of frames from the camera and a set of motor commands. 

The visual information comprises five consecutive images with $128 \times 128$ size. The motor command comprises a vector of 12 motor angle values. To merge action commands with the images, we designed two encoders responsible for each type of observation: a Visual encoder and an action encode. The visual encoder consisted of 48-convolutional layers \cite{o2015introduction, he2016deep}, followed by five Long Short-Term Memory (LSTM) modules \cite{hochreiter1996lstm}, and an action encoder consisting of three fully connected layers with a Leaky Rectified Linear Unit.

The Action Encoder maps the action vectors into some high-dimensional latent space through the fully-connected layer networks. The Visual encoder extracts information using a convolutional neural network and leverages LSTM units to process the features and embed the concept of time, allowing the model to learn the temporal dynamics within the sequential pictures \cite{oh2018automated}. Then, both outputs are concatenated and passed through five fully connected layers. Each layer is followed by a layer of Leaky ReLU activation function, and the output comprises six numbers representing the robot future change in position and orientation.

During deployment, the robot proposes candidate motor commands. Each set of motor commands is fused with five recent gray-scale images to form the input data. Since the egocentric visual self-model runs in parallel on the GPU, we can obtain future states of the robot corresponding to multiple sets of candidate actions in a single cycle. This process is like a robot imagining its own future states through different trials. The robot then selects the action most suited for its objective functions. Therefore, a planner needs enough action candidates to allow the model to select the optimal action. The maximum number of motor commands per model run is the batch size on the GPU. The visual self-model pipeline ran at about 5 Hz during our real-world experiments. We tested 100 candidate motor commands in parallel on an NVIDIA Geforce RTX 3090.

\subsection{Training Egocentric Visual Self-model}
The egocentric visual self-model was trained in a simulation environment using a rug-like texture similar to the real world, along with three additional textures (Figure in supplementary materials). The rug texture was chosen for its rich features, which provide the Visual Encoder with valuable information for optical flow estimation. To prevent overfitting to a single texture, the imported terrain and texture maps were modified (through rotation and scaling) for each training episode. Before training the model, we normalized the ground-truth value to the same standard deviation since the output values contain positive and negative numbers and different ranges. This pre-process effectively avoids model bias, which may predict specific output values better than others that do not significantly impact the loss. We use an Adam optimizer with a learning rate of 1e-4 at the beginning and would be scaled by 0.1 if the validation loss does not decrease over twenty epochs. The loss function can be written as:$ L = MSE(f_p(I_t, A_t) - S_{t+1})$. We have seven different environment domains, as shown in the supplementary material. To reduce the gap between the simulation and the real-world environment, we use the same outer morphology of the main body CAD of our physical robot. Other robot CAD parts are simplified when building our URDF files to speed up the simulation computation. Also, we measure the mass of all the components and set the same number in the file. Besides, we calibrate the number of simulation timestamps in each step by running the same motor trajectories in the real world and the simulation. The trajectories are hand-designed to move each joint from a lower limited position to a limited upper position. We set 60 simulated steps for each action command to ensure the motor commands can be fully executed in the simulation. We collect 100 thousand steps for each texture in the simulation to train the visual-self model.

\begin{figure}
    \includegraphics[width=0.49\textwidth]{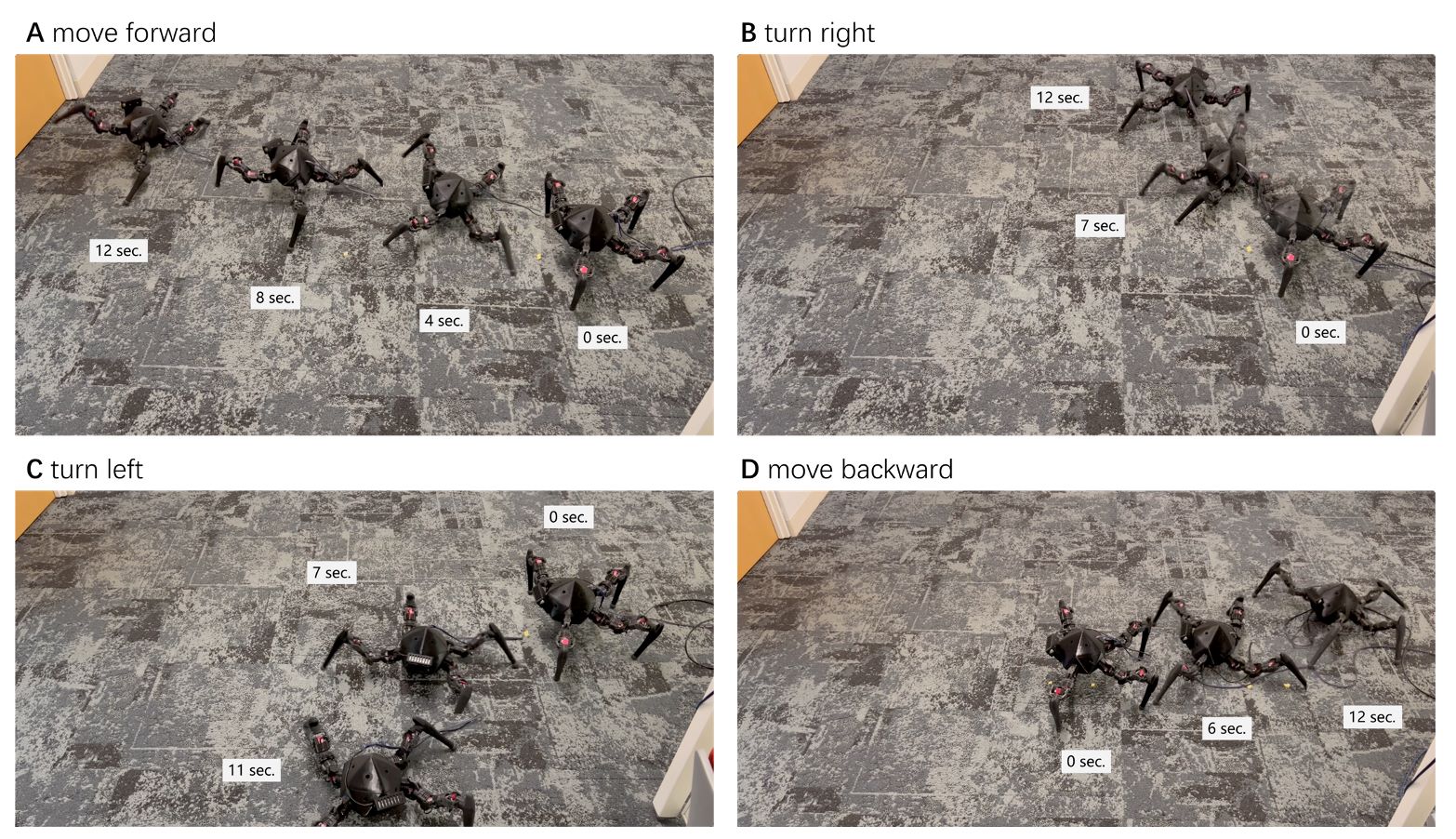}
    \caption{Basic locomotion tasks in the real-world environment. We deployed an egocentric visual self-model on a legged robot. The robot only relies on image sequences from the front camera and action commands to achieve (A) moving forward, (B) turning right, (C) turning left, and even (D) moving backward. It learns the skill from the simulation to anticipate one hundred possible future states by perceiving the latest visual information and motor commands it will actuate.}
    \label{fig:realworld_environment_tasks}
    \vspace{-10pt}
\end{figure}

\section{Experiments}

Our evaluation includes various aspects of the implemented egocentric visual self-model, focusing on its ability to predict future robot states, adaptability to different robots or unseen terrains, and resilience in situations involving hardware anomalies.

\subsection{Real-world Experiments and Robustness in Unseen Environments}
We conducted real-world experiments to assess the performance and robustness of our egocentric visual self-model in various locomotion tasks and unseen environments (Fig. \ref{fig:realworld_environment_tasks}). The robot, equipped with an RGB camera, successfully navigated and maintained stable trajectories while moving forward, turning right, turning left, and moving backward. We compared our method with several baselines and variations, including a Sinusoidal Gait, a Gait Generator, and a version of our method that uses IMU data instead of visual input.

\begin{figure}
    \centering
    \includegraphics[width=0.47\textwidth]{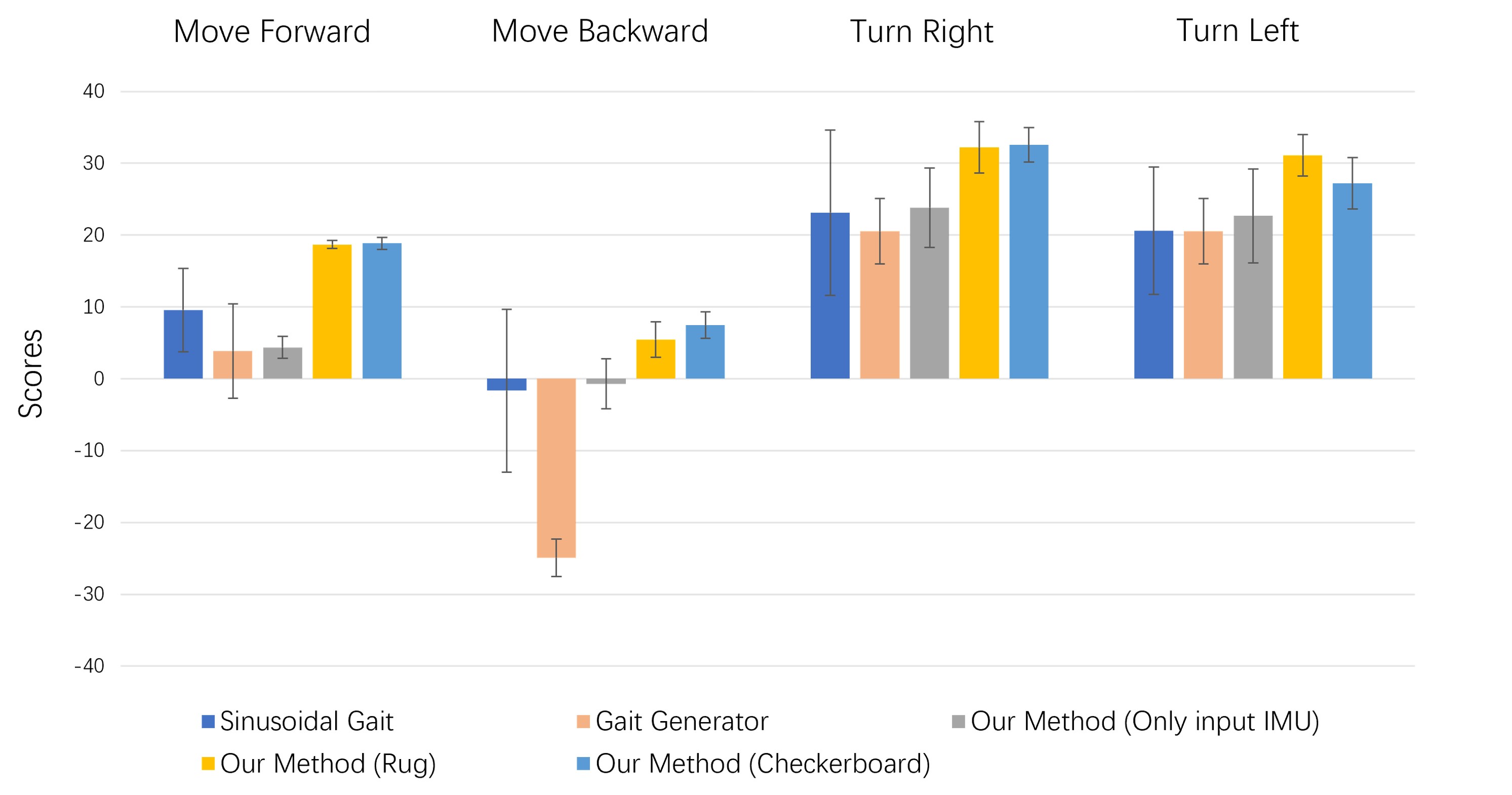}
    \caption{Evaluate the egocentric visual self-model in the real world with four baselines (n=3).}
    \label{fig:realworld_evaluation}
    \vspace{-10pt}
\end{figure}

To evaluate the robustness of our self-model in unseen environments and varying visual conditions, we tested our method on a rug texture used during training in simulation and a checkerboard texture that the robot had never encountered before (Fig. \ref{fig:realworld_evaluation}). Additionally, we conducted experiments on a carpet littered with slippery paper scraps (Fig. \ref{fig:robustness}A-F) to assess the model's performance in challenging terrains. Despite the novel texture and the slippery surface causing slippage, our model demonstrated resilience and maintained stable locomotion.

Figure\ref{fig:realworld_evaluation} presents the mean scores by the robot for each method across three trials. Our method outperforms all other baselines and variations, highlighting the effectiveness of leveraging egocentric visual information for robot self-modeling and control in real-world scenarios. Notably, our method maintains strong performance even when tested on the unseen checkerboard texture, demonstrating its ability to generalize to novel visual environments.

\subsection{Evaluating the Contribution of the Visual Encoder}

To quantitatively evaluate the effectiveness of the Visual Encoder in our pipeline, we compared our method with two baselines. First, we compared our method with the "Commands only - no vision" baseline. The goal of this baseline is to remove the contribution of images from the model and rely only on motor commands to predict future states, equivalent to open-loop control. Intuitively, it represents animals or humans learning to walk blindly without visual input. This baseline model learns the robot's forward-kinematic body model through data-driven like previous work \cite{kwiatkowski2019task}. Therefore, by contrasting our approach against this baseline, we aim to emphasize the importance of visual information. The second baseline replaces image sequences with explicit IMU data as input and remains consistent with our model in the motor action module. We use this baseline to compare the contribution of the high-dimensional visual information over the low-dimensional IMU information typically used in robotics. In the real-world experiment, we used high-accuracy IMU (HWT905-TTL MPU-9250 9-axis Gyroscope+Angle (XY 0.05° error).

We execute ten episodes for each task in each environment, containing 56 steps. Table\ref{tab:quantitive} shows the results of our experiments in the simulation. We test our model with four terrain textures in supplementary materials. We compare our method to the baselines using mean prediction loss and evaluation metrics for each task. The results indicate our approach is better than any other baselines in four tasks across all terrains. We believe that high-dimensional observation should conclude the information from IMU data. Visual observation can also provide additional information about robot configuration, kinematics, and dynamics. Thus, the experiment results demonstrate the importance of visual information in predicting robot dynamics and kinematics.

\begin{table}[!ht]
    \centering
    \caption{Quantitative evaluations}
    \scalebox{0.7}{%
    \begin{tabular}{|c|c|c|c|c|c|}
    \hline
        ~ & ~ & Move Forward & Move Backward & Turn Right & Turn Left \\ \hline
        Action Only & Mean & 3.10E-03 & 1.90E-03 & 2.80E-03 & 2.30E-03 \\ \hline
        ~ & Std & 3.90E-03 & 1.90E-03 & 3.80E-03 & 2.40E-03 \\ \hline
        IMU Only & Mean & 2.40E-03 & 1.80E-03 & 2.10E-03 & 2.20E-03 \\ \hline
        ~ & Std & 3.00E-03 & 1.90E-03 & 1.80E-03 & 2.50E-03 \\ \hline
        Ours (rug) & Mean & 1.90E-03 & 1.40E-03 & 1.60E-03 & 1.50E-03 \\ \hline
        ~ & Std & 2.30E-03 & 1.70E-03 & 1.70E-03 & 1.60E-03 \\ \hline
        Ours (grid) & Mean & 1.90E-03 & 1.40E-03 & 1.60E-03 & 1.50E-03 \\ \hline
        ~ & Std & 2.30E-03 & 1.70E-03 & 1.70E-03 & 1.60E-03 \\ \hline
        Ours (color Dots) & Mean & 1.40E-03 & 1.30E-03 & 1.60E-03 & 1.40E-03 \\ \hline
        ~ & Std & 2.20E-03 & 1.60E-03 & 1.50E-03 & 1.30E-03 \\ \hline
        Ours (grass) & Mean & 1.40E-03 & 1.30E-03 & 1.60E-03 & 1.30E-03 \\ \hline
          & Std  & 1.50E-03  & 1.40E-03  & 1.60E-03  & 1.40E-03  \\ \hline
    \end{tabular}}
    \label{tab:quantitive}
    \vspace{-12pt}
\end{table}

\begin{figure}
    \centering
    \includegraphics[width=0.49\textwidth]{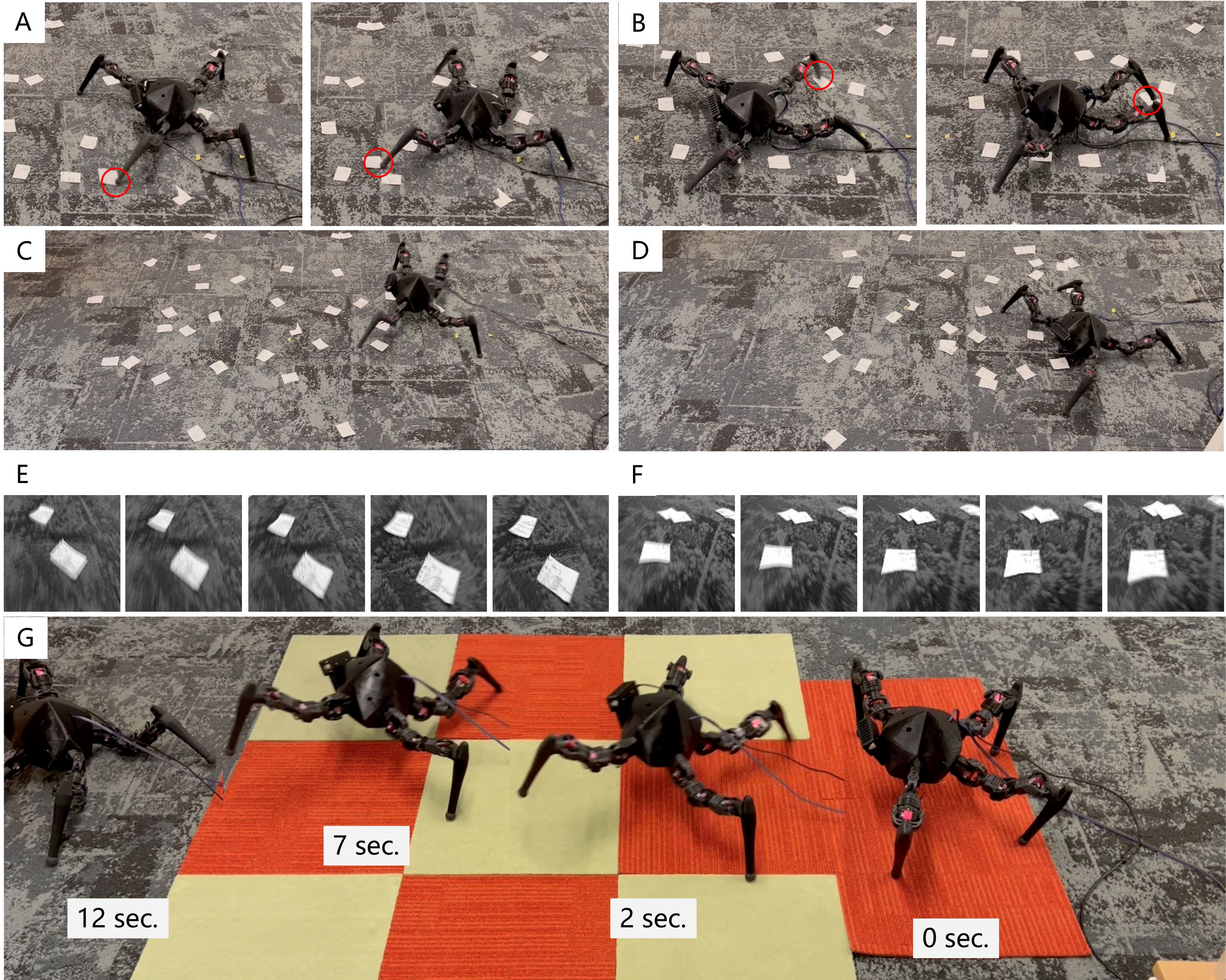}
    \caption{Assessing egocentric visual self-models on unseen terrain. (A, B, C, D and E) We performed the same experiments on carpets with pieces of slippery paper and terrain with checkerboard textures. (C) The robot is moving forward, and its first-view pictures are shown in E. (D) The robot is turning right and its first-view pictures are shown in F. (G) The robot moves forward with an egocentric visual self-model on unseen terrain. }
    \label{fig:robustness}
    \vspace{-10pt}
\end{figure}

Four different terrains were used during the training phase. We also tested our robot on a checkerboard texture, which was not seen during the training, as shown in Fig.\ref{fig:robustness}. These experiments underscore the adaptability of our approach to new environments and its robustness to visual perturbations.

\subsection{Generalizability and Transfer Learning}

In this section, we evaluate the generalizability of the visual encoder in our model by conducting experiments on three additional robots in the Pybullet simulation environment. The main goal is to determine whether the pre-trained visual encoder from robot 0 can be used to train the egocentric visual self-models of the new robots while maintaining performance and requiring less data during the training process. We designed two baselines and conducted quantitative evaluations to demonstrate the generalizability of our model.

\begin{figure*}[h]
    \centering
    \includegraphics[width=0.95\textwidth]{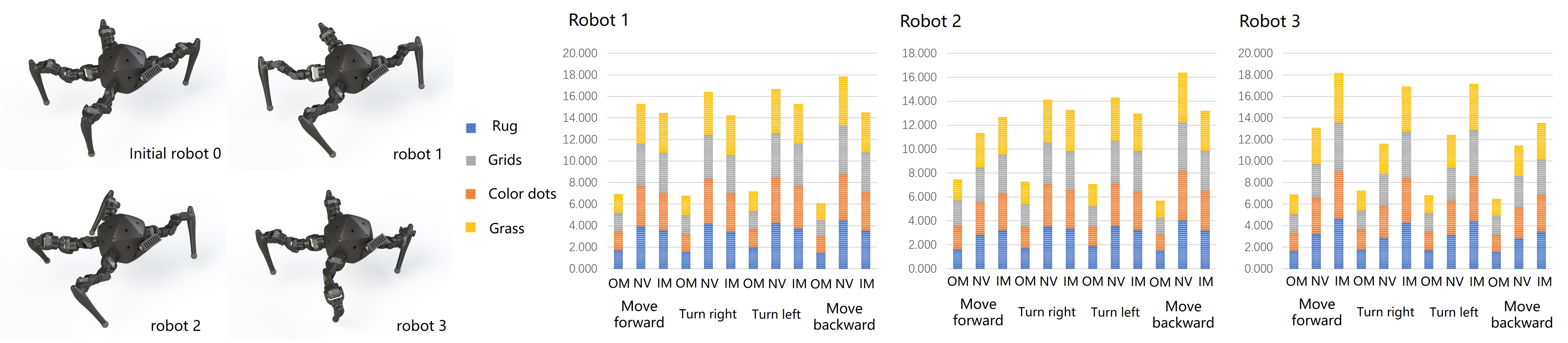}
    \caption{New robot configurations and quantitative evaluations in various terrains and locomotion tasks. Robot 0 is modified to create three new configurations, robot 1, robot 2, and robot 3, by altering leg-body connection orientations, resulting in significant kinematic changes due to the serial connection of leg motors, thus providing a diverse set of test cases for our visual encoder's generalizability. The lower plots show prediction errors for our method (OM), a non-visual method (NV), and the initial model (IM). Our method consistently outperforms the other methods, exhibiting the lowest errors across all terrains and tasks for the three new robots.}
    \label{fig:generalizability_eval2}
    \vspace{-10pt}
\end{figure*}

To assess the generalizability of our model's visual encoder, we selected three different robots in the Pybullet simulation environment, as shown in Fig.\ref{fig:generalizability_eval2}. The experiment of our Robot 3(with the second and third joints of each leg being rotated by 90 degrees) has significant kinematic changes. These robots possess varying morphologies and complexities, ensuring a diverse and challenging set of test cases for our model. The pretrained visual encoder from robot 0 was used as the starting point for training the egocentric visual self-models of the new robots.

During the training process, the visual encoder weights were kept frozen, meaning that the model did not learn any new visual features specific to the new robots. Instead, it relied on the visual features learned from the initial robot. We used less data (200k steps) for training the new egocentric visual self-models compared to the data used for training the initial robot 0 (400k steps). 

We designed two baselines for quantitative evaluation. The first baseline, a non-visual method (NV), relies only on non-visual information for state prediction and control. The second baseline (IM) utilized the initial egocentric visual self-model of robot 0 without additional training for the new robots. The comparison between our method and the IM baseline highlights the substantial reduction in prediction errors achieved by fine-tuning the egocentric visual self-model with a small batch of data for each new robot. The pre-trained visual encoder's ability to capture crucial information about the robot's state, combined with the fine-tuning of the self-model, enables efficient adaptation to new robot configurations while significantly improving prediction accuracy.

The results (Fig.\ref{fig:generalizability_eval2}) demonstrate that our method (OM) consistently outperformed both baselines, exhibiting the lowest prediction errors across all terrains and tasks for the three new robots. These results highlight the visual encoder's ability to capture essential information about robot dynamics and facilitate efficient learning and adaptation for different robotic systems.

\subsection{Applicability to a Humanoid Robot}
\begin{figure}[h]
    \centering
    \includegraphics[width=0.47\textwidth]{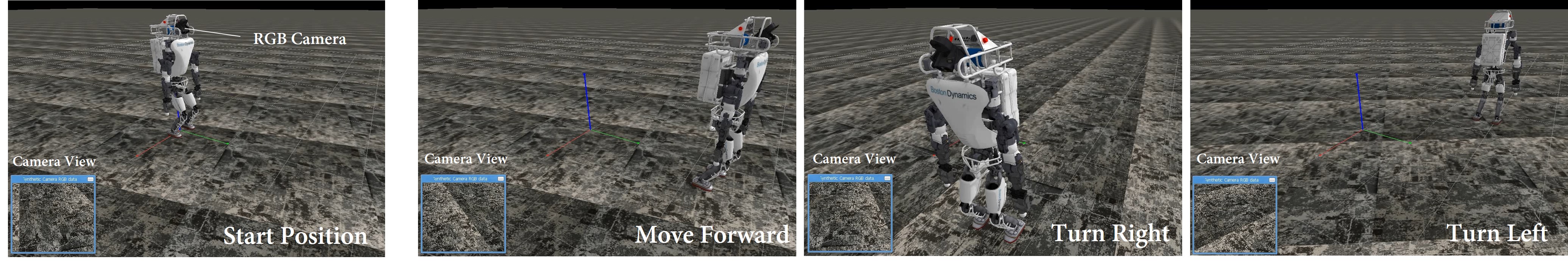}
    \caption{Egocentric visual self-model deployed on the humanoid robot Atlas in simulation. The Atlas robot successfully performs basic locomotion tasks such as moving forward, turning right, and turning left, demonstrating the adaptability and generalizability of our approach to complex robotic systems with vastly different morphologies.}
    \label{fig:atlas}
    \vspace{-12pt}
\end{figure}

To further demonstrate the versatility and adaptability of our egocentric visual self-model, we conducted an additional experiment on a significantly different robot: the humanoid robot Atlas. Controlling such a sophisticated system is challenging due to its complex morphology and dynamics. To focus on testing the basic locomotion capabilities based on our egocentric visual self-model and its adaptability to different robots, we froze the upper body joints and used only 6 degrees of freedom, including the hip, knee, and ankle joints. We evaluated the performance of our method on the Atlas robot in a simulated environment. The results, shown in Figure\ref{fig:atlas}, demonstrate that our method can be successfully deployed on this robot, enabling it to move forward and turn left and right. This highlights the generalizability of our approach to robots with vastly different morphologies and complexity compared to the initial robot used in our experiments.

We also conducted a quantitative evaluation of our method's performance on the Atlas robot. First, we assessed the prediction errors of our egocentric visual self-model compared to a baseline that does not use visual input (Table \ref{tab:atlas_pred}). Our method significantly outperformed the baseline, achieving much lower prediction errors across all locomotion tasks (forward, right, and left). Furthermore, we compared the locomotion performance of our method against the baseline without visual input and a gait generator (Table \ref{tab:atlas_performance}). Our method consistently outperformed both baselines in terms of the mean distance traveled in each direction (forward, right, and left), demonstrating its effectiveness in enabling stable and efficient locomotion for the Atlas robot.

These experiments further validate the adaptability and generalizability of our egocentric visual self-model to various robotic systems, including humanoid robots with complex morphologies and dynamics. The successful deployment of our method on the Atlas robot highlights its potential as a universal tool for enabling basic locomotion capabilities in diverse robotic platforms, paving the way for more advanced and agile behaviors in the future.

\begin{table}[h]
\centering
\caption{Humanoid Robot State Prediction Errors}
\begin{tabular}{|c|cc|cc|}
\hline
                 & \multicolumn{2}{c|}{\textbf{Our Method}}          & \multicolumn{2}{c|}{\textbf{No Images Input Baseline}} \\ \hline
                 & \multicolumn{1}{c|}{\textbf{MEAN}} & \textbf{STD} & \multicolumn{1}{c|}{\textbf{MEAN}}    & \textbf{STD}   \\ \hline
\textbf{Forward} & \multicolumn{1}{c|}{5.99E-04}      & 3.16E-05     & \multicolumn{1}{c|}{3.38E-01}         & 2.16E-01       \\ \hline
\textbf{Right}   & \multicolumn{1}{c|}{7.05E-04}      & 2.24E-05     & \multicolumn{1}{c|}{4.61E-01}         & 1.49E-01       \\ \hline
\textbf{Left}    & \multicolumn{1}{c|}{5.77E-04}      & 4.04E-05     & \multicolumn{1}{c|}{4.71E-01}         & 1.95E-01       \\ \hline
\end{tabular}
\label{tab:atlas_pred}
\vspace{-10pt}
\end{table}

\begin{table}[h]
\centering
\caption{Humanoid Robot Locomotion Performance Comparisons}
\scalebox{0.75}{%
\begin{tabular}{|c|cc|cc|cc|}
\hline
\multicolumn{1}{|l|}{} & \multicolumn{2}{c|}{\textbf{Our   Method}}        & \multicolumn{2}{c|}{\textbf{No   Images Input Baseline}} & \multicolumn{2}{c|}{\textbf{Gait   Generator}}    \\ \hline
\multicolumn{1}{|l|}{} & \multicolumn{1}{c|}{\textbf{MEAN}} & \textbf{STD} & \multicolumn{1}{c|}{\textbf{MEAN}}     & \textbf{STD}    & \multicolumn{1}{c|}{\textbf{MEAN}} & \textbf{STD} \\ \hline
\textbf{Forward}       & \multicolumn{1}{c|}{1.27E+00}      & 1.34E-01     & \multicolumn{1}{c|}{4.06E-01}          & 8.13E-01        & \multicolumn{1}{c|}{4.50E-01}      & 1.39E+00     \\ \hline
\textbf{Right}         & \multicolumn{1}{c|}{2.01E+00}      & 3.16E-01     & \multicolumn{1}{c|}{-6.96E-01}         & 2.08E-01        & \multicolumn{1}{c|}{3.46E-01}      & 1.54E+00     \\ \hline
\textbf{Left}          & \multicolumn{1}{c|}{1.67E+00}      & 1.94E-01     & \multicolumn{1}{c|}{1.53E+00}          & 1.89E-01        & \multicolumn{1}{c|}{4.10E-01}      & 7.95E-01     \\ \hline
\end{tabular}}
\label{tab:atlas_performance}
\vspace{-10pt}
\end{table}

\subsection{Autonomous anomaly identification and adaptation}

\begin{figure}[h]
    \centering
    \includegraphics[width=0.49\textwidth]{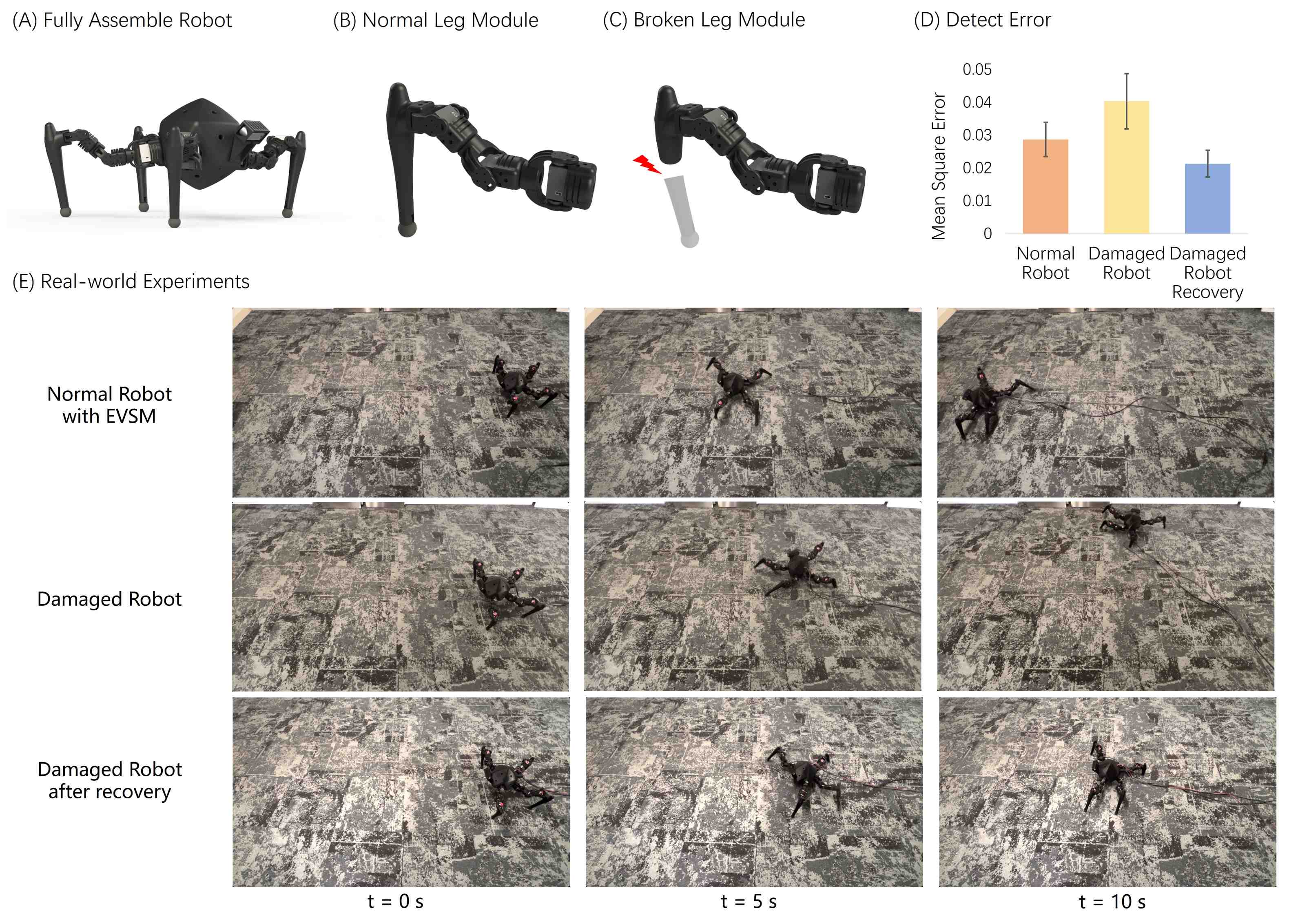}
    \caption{Description of anomalies. (A) Our fully assembled robot has four legs and a camera aimed at the ground. (B) The robot leg module consists of three motors and a rubber ball foot in contact with the ground. (C) To test the resilience ability of the robot, we cut the end-link model in half to act as a damaged robot leg. (D) Error detected by the visual odometry model. (E) Resilience experiments in the real world. The first row shows the normal robot moving forward with the egocentric visual self-model. In the second row, we replaced the right hind leg of the robot with a damaged leg and deployed the same egocentric visual self-model to control the robot. The last row presents the updated egocentric-visual self-model on the damaged robot to control the robot to move forward. }
    \label{fig:rw_resilience}
    \vspace{-10pt}
\end{figure}

The egocentric visual model can endow the robot with abilities to recognize hardware anomalies, like broken legs or damaged actuators, and recover by adapting to the new body configuration. We conducted anomaly-detection experiments using a different leg module, which has a broken end-link as shown in Fig.\ref{fig:rw_resilience} (A-C). Such damage typically happens when subjected to external force impact or fatigue fracture during testing or deployment and is generally difficult to detect using direct sensors. Recovery usually requires human intervention to determine the cause and shut down the robot for repair. 

To detect the anomaly, we first trained a deep visual odometry model using the same data as the egocentric visual self-model. The model takes 7 consecutive pictures captured by the camera between two steps as input and returns the state changes ($\Delta x,\Delta y, \Delta z, \Delta roll, \Delta pitch, \Delta yaw$). This information is not predictive and does not factor in motor commands. The visual odometry captures the current (not the future) state of the robot. The visual odometry model has the same architecture as our egocentric visual self-model model, consisting of the visual encoder and MLPs, but omitting the motor command encoder. 

We deployed the learned visual odometry model directly on the physical robot. As shown in the first two columns of Fig.\ref{fig:rw_resilience}D, the robot can recognize that an abnormal situation (leg broken in this case) may have occurred based on a large discrepancy between predicted and actual motion. The first two rows in Fig.\ref{fig:rw_resilience}E demonstrate the difference when we use the egocentric visual self-model to control the normal robot and the damaged robot to do the same forward motion task. Under normal circumstances, the robot will choose a forward gait according to its predicted future state to achieve forward motion. However, since its right rear leg was broken, following the original gait caused it to deflect to the right. The discrepancy between predicted and actual motion can trigger an alarm. It can also trigger an auto-recovery process as follows.

In order for the robot to regain its previous ability to accurately predict future states based on visual information and motion commands, the damaged robot needs to re-learn its self-model to match its new, altered body configuration. By running motor babbling in the real environment for about 30 minutes, the robot collected 7,000 steps of data. This data includes the motor commands of the robot and the images captured by the camera. The Visual Odometry (VO) model outputs the change of robot state from the visual information as the ground truth label for updating the visual self-model. Egocentric visual self-models can be re-trained with this new data to successfully and accurately predict future states. Once the robot states predicted by the self-model match the actual robot states measured by the Visual Odometry, the self-model retraining has been completed.  Fig.\ref{fig:rw_resilience}D shows how the error becomes lower after updating the model, and the robot can recover to perform forward locomotion in Fig.\ref{fig:rw_resilience}E.

\section{Conclusion}
In this work, we introduce a new approach to robot self-modeling that leverages egocentric visual perception to predict future states conditioned on motor commands. To the best of our knowledge, this is the first demonstration of a legged robot using self-modeling to successfully perform locomotion tasks using only first-person visual observations, without relying on proprioceptive sensors or prior knowledge of the robot's morphology, kinematics, or dynamics.

Through a series of experiments in both simulation and the real world, we showcase the effectiveness of our egocentric visual self-model in enabling basic locomotion, adaptation to unseen terrains, and transfer learning across different robot morphologies. The successful deployment of our approach on the humanoid robot Atlas further highlights its potential for controlling complex robotic systems. Another contribution of our work is the robot's ability to autonomously detect and adapt to hardware anomalies, such as damaged components, by comparing predicted and actual motion using a pre-trained visual odometry model. This resilience is crucial for real-world deployment, where robots may encounter unexpected failures.

Future work could explore the integration of additional sensory modalities, hierarchical planning, and long-term memory to enable more sophisticated behaviors and longer prediction horizons. We believe that our work provides a solid foundation for future research in visual robot self-modeling. By demonstrating the efficacy of learning directly from visual input, we open up new possibilities for giving robots autonomy, adaptability, and resilience combined with vision.

\section*{APPENDIX}
The supplementary materials for this work are available at: https://github.com/H-Y-H-Y-H/Egocentric\_VSM

\bibliographystyle{IEEEtran}
\bibliography{IEEEabrv, ref}

\end{document}